\documentclass[10pt,twocolumn,letterpaper]{article}

\usepackage{iccv}
\usepackage{times}
\usepackage{epsfig}
\usepackage{graphicx}
\usepackage{diagbox} %for labels of table
\usepackage{amsmath}
\usepackage{amssymb}
\usepackage{algorithm}
\usepackage{booktabs}
\usepackage{lipsum}
\usepackage{algorithm}  
\usepackage{algpseudocode}  
\usepackage{amsmath}
\usepackage{multirow}
\usepackage[dvipsnames,svgnames,x11names]{xcolor}
\usepackage{enumitem}
\setitemize{noitemsep,topsep=0pt,parsep=0pt,partopsep=0pt}
\usepackage[accsupp]{axessibility} % Improves PDF readability for those with disabilities.

\usepackage[breaklinks=true,bookmarks=false]{hyperref}
\hypersetup{
    colorlinks=true,
    citecolor=Teal,
}

\iccvfinalcopy % *** Uncomment this line for the final submission

\def\iccvPaperID{8235} % *** Enter the ICCV Paper ID here

\def\ie{\emph{i.e.}}

\def\etal{\emph{et al.}}

\newcommand*\samethanks[1][\value{footnote}]{\footnotemark[#1]}

\ificcvfinal\fi

\begin{document}

%%%%%%%%% TITLE
\title{Lipschitz Continuity Guided Knowledge Distillation}

% \and

\author{
  \textbf{Yuzhang Shang$^1$\thanks{\, Equal contribution. $\dag$\,  Corresponding author.}, Bin Duan$^1$\samethanks{}, Ziliang Zong$^{2}$,}
  \textbf{Liqiang Nie$^{3}$, Yan Yan$^{1\dag}$}\vspace{3pt}\\
  $^{1}$Department of Computer Science, Illinois Institute of Technology, USA \\ 
  $^{2}$Department of Computer Science, Texas State University, USA \\
  $^{3}$School of Computer Science and Technology, Shandong University, China \\ 
  \tt\small{\{yshang4, bduan2\}@hawk.iit.edu, ziliang@txstate.edu}\vspace{-2pt}\\
  \tt\small{nieliqiang@gmail.com, yyan34@iit.edu} 
}

% \footnote{$\star$ stands for the corresponding author}
% \name{Yuzhang Shang$^{\star}$$^{\dagger}$ \qquad Bin Duan$^{\dagger}$ \qquad Ziliang Zong$^{\dagger}$ \qquad Liqiang Nie$^{\star}$ \qquad Yan Yan$^{\dagger}$}

% \address{$^{\star}$ School of Computer Science and Technology, Shandong University, China\\
%       $^{\dagger}$ Department of Computer Science, Texas State University, USA }

\maketitle
% \newcommand\blfootnote[1]{%
% \begingroup
% \renewcommand\thefootnote{}\footnote{#1}%
% \addtocounter{footnote}{-1}%
% \endgroup
% }

\ificcvfinal\thispagestyle{empty}\fi

% \blfootnote{$\dagger$ Equal contribution, $\star$ Corresponding author.}

\begin{abstract}
Knowledge distillation has become one of the most important model compression techniques by distilling knowledge from larger teacher networks to smaller student ones. Although great success has been achieved by prior distillation methods via delicately designing various types of knowledge, they overlook the functional properties of neural networks, which makes the process of applying those techniques to new tasks unreliable and non-trivial. To alleviate such problem, in this paper, we initially leverage Lipschitz continuity to better represent the functional characteristic of neural networks and guide the knowledge distillation process. In particular, we propose a novel Lipschitz Continuity Guided Knowledge Distillation framework to faithfully distill knowledge by minimizing the distance between two neural networks' Lipschitz constants, which enables teacher networks to better regularize student networks and improve the corresponding performance. We derive an explainable approximation algorithm with an explicit theoretical derivation to address the NP-hard problem of calculating the Lipschitz constant. Experimental results have shown that our method outperforms other benchmarks over several knowledge distillation tasks (e.g., classification, segmentation and object detection) on CIFAR-100, ImageNet, and PASCAL VOC datasets. Our code is available at \url{https://github.com/42Shawn/LONDON/tree/master}.
\end{abstract}
\section{Introduction}
Recently, deep learning models have driven great advances in computer vision~\cite{he2016deep,girshick2015fast}, natural language process~\cite{sutskever2014sequence,pennington2014glove}, information retrieval~\cite{wei2019neural,MMGCN} and multi-modal modelling~\cite{hu2021coarse, hu2021video}. To meet the buoyant demand of equipping those cumbersome models in resource-constrained edge devices, researchers have proposed several network compression paradigms, such as network pruning~\cite{lecun1989optimal,han2015deep}, network quantization~\cite{hubara2016binarized} and knowledge distillation (KD)~\cite{hinton2015distilling}. Among these compression methods, KD helps the
training process of a smaller network (student) by transferring knowledge from a larger one (teacher). As one of the first innovators, Hinton \etal~\cite{hinton2015distilling} proposed using soft labels of the larger networks to supervise the training process of the smaller ones. These soft labels are usually interpreted as a form of unseen knowledge distilled from teachers.

Apart from treating soft labels as distilled knowledge, various kinds of knowledge are designed in~\cite{yim2017gift,heo2019comprehensive,tian2019contrastive,wang2019pay}. For example, Romero \etal~\cite{romero2014fitnets} presented to train intermediate layers of students with guidance of the corresponding layers of teachers, which initiates the subsequent flourishing studies on feature-based knowledge distillation. Researchers~\cite{yim2017gift,lee2018self,tung2019similarity} also modulated the relations among adjacent feature maps as additional knowledge to assist the training of student networks. Unfortunately, most of these feature-based KD methods solely focus on aligning the shallow information but overlook the high-level information of both networks, \ie, the students mechanically mimicking teachers' actions while neglecting their interior qualities. Thereby, previous studies consider networks as black-boxes and heuristically select features without any functional properties \cite{tian2019contrastive, wang2019pay, zagoruyko2016paying}, which impedes a universal representative of knowledge to be distilled. To address this problem, we argue that leveraging networks' functional properties to derive high-level knowledge is able to strengthen the performance of KD. 

In this paper, we incorporate Lipschitz continuity into KD, considering neural networks as functions rather than black-boxes. By definition in Eq.~\ref{definition:lip}, Lipschitz constant\footnote{The Lipschitz constant of a function $\Vert f\Vert_{Lip}$ is the maximum norm of its gradient in the domain set, which reflects Lipschitz continuity of the function.} is the upper bound of the relationship between input perturbation and output variation for a given distance, representing the robustness and expressiveness of neural networks \cite{bartlett2017spectrally, miyato2018spectral, lyu2020autoshufflenet}. Specifically, authors in \cite{ miyato2018spectral, yoshida2017spectral} demonstrated the effectiveness of the Lipschitz constant by constraining the weights of the discriminator in a generative adversarial network (GAN). Besides, many studies in representation learning \cite{bengio2013representation, tian2019multimodal} demonstrate that deep neural networks are competent in learning high-level information with increasing abstraction. Inspired by this, we devise a scheme to capture the Lipschitz continuity (\ie, calculate the Lipschitz constant for every intermediate block) of the teacher networks and adopt the captured continuity as knowledge to guide the training of student networks. It is worth noting that Lipschitz constant computation is a NP-hard problem \cite{virmaux2018lipschitz}. We address this problem by proposing an approximation algorithm with a tight upper bound. In particular, we design a Transmitting Matrix ($\mathbf{TM}$) for each block and calculate the spectral norm of $\mathbf{TM}$ through an adopted iteration method to avoid the high complexity of learning large intermediate matrices. We then aggregate all Lipschitz constants calculated from $\mathbf{TM}$s as the knowledge of the Lipschitz continuity that are transferred to student networks. Importantly, Lipschitz continuity loss function is backpropagation-friendly for training deep networks because of its differentiability. 

Overall, the contributions of this paper are four-fold:
\begin{itemize}[leftmargin=*]
    \item To the best of our knowledge, we are the first on utilizing a high-level functional property, Lipschitz continuity in knowledge distillation, to supervise student networks' training process. In addition, we theoretically explain the effectiveness of our method from the perspective of network regularization and then empirically consolidate this explanation.
    \item We propose a novel knowledge distillation framework, \textbf{L}ipschitz c\textbf{ON}tinuity Guided Knowledge  \textbf{D}istillati\textbf{ON} (\textbf{LONDON}) for distilling knowledge from the Lipschitz constant.
    \item To avoid the NP-hard Lipschitz constant calculation, we devise a Transmitting Matrix to numerically approximate the Lipschitz constant of networks in the KD process.
    \item We perform experiments on different knowledge distillation tasks such as classification, object detection, and segmentation. Our proposed method achieves the state-of-the-art results in these tasks on CIFAR-100, ImageNet, and VOC datasets.
\end{itemize}

\section{Related Work}
\noindent \textbf{Lipschitz Continuity and Spectral Norm of Neural Network.} The study of adversarial machine learning \cite{kurakin2016adversarial, papernot2016transferability} shows that neural networks are highly vulnerable to attacks based on small modifications of the input to the model at test time, and estimating the regularity of such architectures is essential for practical applications and generalization improvement. Previous efforts~\cite{virmaux2018lipschitz, miyato2018spectral,neyshabur2017exploring} have studied one of the critical characteristics to assess the regularity of deep networks: the Lipschitz continuity of deep learning architectures. 

Lipschitz constants, which upper bound the relationship between input perturbation and output variation for a given distance, are introduced to secure the robustness of neural networks to small perturbations. This Lipschitz constant $\Vert f \Vert_{Lip}$ can be seen as a norm to measure the function's degree of Lipschitz continuity.
Apart from some theoretical studies \cite{bartlett2017spectrally,luxburg2004distance,neyshabur2017exploring} explaining that novel generalization bounds critically rely on the Lipschitz constant of the neural network, Lipschitz continuity of neural networks is widely studied for achieving the state-of-the-art performance in many deep learning topics: 
(i) In image synthesis~\cite{miyato2018spectral,yoshida2017spectral}, researchers used spectral normalization on each layer, an optional approach to constrain the Lipschitz constant of the discriminator for training a GAN on ImageNet, like a regularization term to smooth the discriminator function. And (ii) in adversarial attack machine learning~\cite{weng2018evaluating}, authors propose constraining local Lipschitz constants of neural networks to avoid adversarial attacks. 

Aforementioned efforts underline the significance of Lipschitz constant in neural networks' expressiveness and robustness. Particularly, deliberately constraining Lipschitz continuity (constant) in an appropriate range is proven to be a powerful technique for smoothing networks, which can enhance the model's robustness. On account of this, Lipschitz constant, the functional information of neural networks should be introduced into knowledge distillation model for regularizing the training of student networks.

\noindent \textbf{Knowledge Distillation.} Apart from the seminal design of soft labels~\cite{hinton2015distilling}, the alignment of intermediate feature maps is also transferred as knowledge to student networks~\cite{romero2014fitnets}. Researchers continued digging into feature-based outputs and proposed various designs of feature maps' transformation and combination to define the feature-based knowledge, which largely promotes the performance of KD. For example, Heo \etal~\cite{heo2019knowledge, heo2019comprehensive} designated an activation boundary of hidden neurons in different positions of networks as knowledge for distillation. In \cite{yim2017gift}, Gram matrixes of neural networks' adjacent feature maps, representing the relation between intermediate layers, are also adopted as a form of knowledge. Authors~\cite{lee2018self, chen2020learning, tung2019similarity} constructed a similarity measurement for feature representations using singular value decomposition (SVD) to elicit relations between different layers as transferred knowledge. 

Inspired by those ideas, many methods are proposed for precisely capturing feature-wise knowledge by artlessly piling up complicated mechanisms on knowledge distillation model. For instance, Wang \etal~\cite{wang2019pay} introduced an attention mechanism to assign weights to different CNNs' channels for dynamically determining the critical features to distill. Furthermore, Tian \etal~\cite{tian2019contrastive} introduced contrastive learning to capture correlations and higher-order output dependencies for supervising student network training. This dynamically aligned knowledge almost fully explores potential of distilling networks' output information for supervision.

However, all those feature-based knowledge distillation methods treat neural networks as black-boxes, which are deficient in exploring the functional properties of neural networks via capturing the high-level information. This limitation hinders the applicability and impedes performance improvement. To alleviate the limitation, we introduce Lipschitz continuity to knowledge distillation.

\section{Method}
In this section, we introduce our proposed knowledge distillation framework. We only elaborate the key derivations in this section due to the limited space. Detailed discussions and technical theorems can be found in the supplemental materials. Here, we focus on capturing the functional property of neural networks as knowledge and transferring it in our distillation method in a numerically accessible way.

\begin{figure*}[t]
    \centering
    \includegraphics[width=0.97\textwidth]{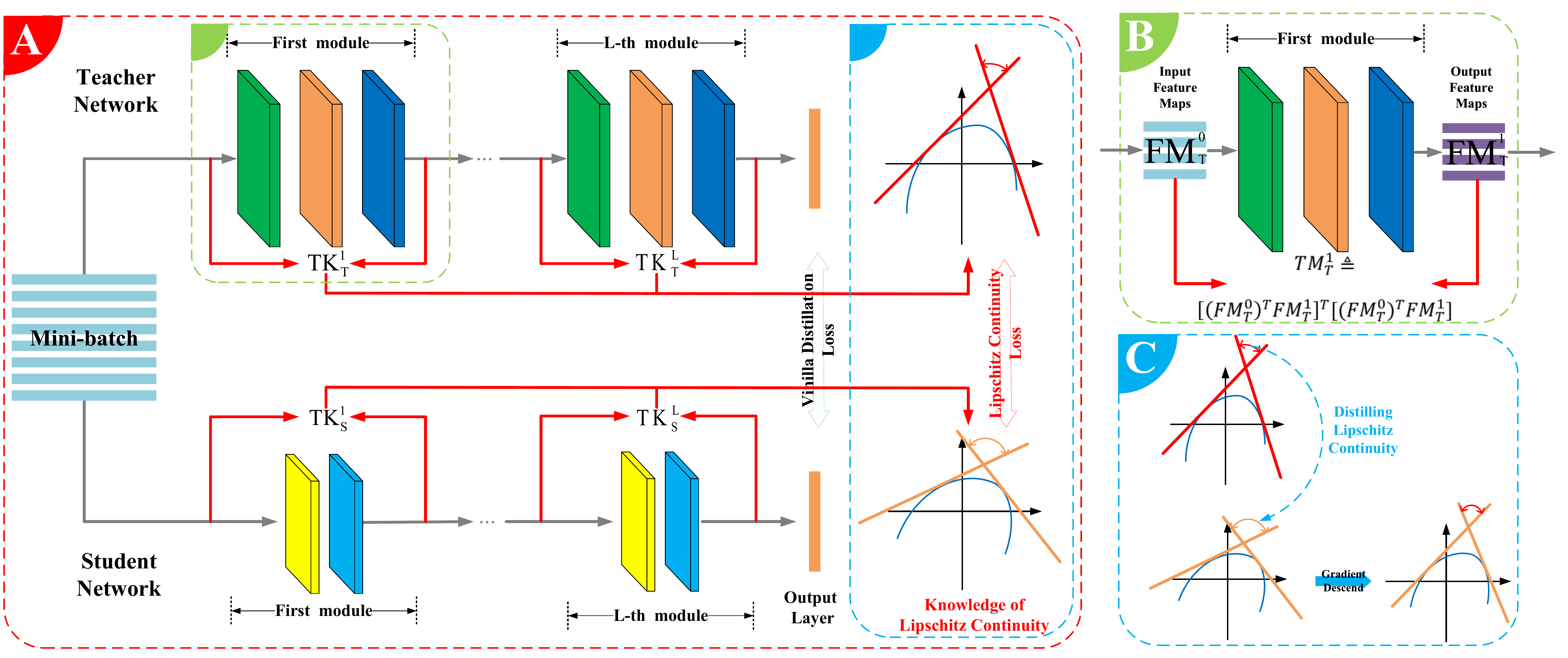}
    \caption{An overview of our proposed LONDON is indicated in \textbf{A}. For the teacher-student backbone, besides the traditional knowledge distillation loss, our proposed Lipschitz continuity distillation loss is the key element. The input and output feature maps of each module are used to format the Transmitting Matrix $\mathbf{TM}^k$ for approximating the module's Spectral Norm as demonstrated in \textbf{B}. Those Spectral Norms are combined to calculate the Lipschitz constant of networks for further distillation via our designed Lipschitz continuity loss function $\mathcal{L}_{Lip}$, which regularizes the student training in a high level showed in \textbf{C}.}
    \label{fig:pipeline}
    \vspace{-0.5cm}
\end{figure*}

\subsection{Preliminary}

We first define a fully-connected neural network with $L$ layers of widths $d_1,\cdots d_L (d = \sum_{k=1}^L d_k)$ as the form of function $f: \mathbb{R}^{d_0} \longmapsto \mathbb{R}^{d_L}$:
\begin{equation}
    f(\mathbf{x}) = (T^{L}\circ\sigma\circ T^{L-1}\circ \cdot\cdot\cdot \circ\sigma\circ T^{1})(\mathbf{x}),
    \label{eq:1}
\end{equation}
where each $T^{(k)}: \mathbb{R}^{d_{k-1}} \longmapsto \mathbb{R}^{d_{k}}$ is an affine function ($d_0$ and $d_L$ are the sizes of network's input and output feature maps) and $\sigma$ performs element-wise activation for feature maps. For $k$-th layer of the networks, $T^{(k)}(\mathbf{u}) = \mathbf{W}^{k}\mathbf{u}+\mathbf{b}^{k}$, where $\mathbf{W}^{k}$ and $\mathbf{b}^{k}$ stand for the weight matrix and bias vector, respectively. For generality purpose, we discard the bias term of the network, so that the network can be simplified as:
\begin{equation}
    f(\mathbf{W}^1,\cdots,\mathbf{W}^L;\mathbf{x}) = (\mathbf{W}^{L}\circ\sigma\circ \mathbf{W}^{L-1}\circ \cdot\cdot\cdot \circ\sigma\circ \mathbf{W}^{1})(\mathbf{x}).
    \label{eq:2}
\end{equation}
Notably, it is sufficient to consider networks with the most straightforward fully-connected layers, since layer with complex structures such as convolution layer can also be denoted as the form of matrix multiplication. We consider a convolution layer with $i$ input channels and $o$ output channels, and the size of the kernel is $w\times h$, resulting in $iowh$ parameters. We can re-arrange the parameters to a matrix of size $o\times ihw$, such that this convolution layer can also be processed in the same way as the other fully-connected layers do. Hence, our analysis has no loss for generality in this configuration of function $f$.

Following Eq.~\ref{eq:2}, we define the function form of the teacher network as $f_T(\mathbf{W}_T^1,\cdots,\mathbf{W}_T^{L_T};\mathbf{x})$, and the student network as $f_S(\mathbf{W}_S^1,\cdots,\mathbf{W}_T^{L_S};\mathbf{x})$, such that the feature-based KD paradigm can be interpreted as: 
\begin{equation}
    \!\!\!\!\forall \mathbf{x} \in  \mathbf{Data},  \mathop{\arg\min}_{\mathbf{W}_S^1,\cdots,\mathbf{W}_S^{L_S}}  Dist( \mathcal{T}(f_T(\mathbf{x})), \mathcal{T}(f_S(\mathbf{x}))),
\end{equation}
where given the same data, the ultimate goal of KD paradigm is to minimize the distance between teacher and student for optimizing the latter's parameters $\{\mathbf{W}_S^{i}\}$. Particularly, $Dist()$ is a distance function, and $\mathcal{T}()$ is certain transformation approach to turn feature maps into more measurable and learnable knowledge. By utilizing those designed knowledge, the student network is forced to mimic the teacher network and hopefully obtains comparable performance with lighter architecture.

Here, we introduce Lipschitz Continuity into KD paradigm as universal information of neural networks based on the functional property of networks. To make Lipschitz constant calculation numerically feasible, we further propose an approximation for the Lipschitz constant and use power iteration method to calculate this approximation.

\subsection{The functional information of neural networks: Lipschitz Continuity}

\noindent\textbf{Definition 1.} A function $f : \mathbb{R}^{n} \longmapsto \mathbb{R}^{m}$ is called Lipschitz continuous if there exists a constant $L$ such that:
\begin{equation}
    \forall \mathbf{x,y}  \in  \mathbb{R}^{n}, \Vert f(\mathbf{x}) - f(\mathbf{y})\Vert_2 \leq L\Vert \mathbf{x} - \mathbf{y}\Vert_2.
    \label{definition:lip}
\end{equation}
The smallest $L$ that can hold the inequality is called Lipschitz constant of function $f$, denoted as $\Vert f \Vert_{Lip}$. 
By Definition 1, $\Vert f \Vert_{Lip}$ has an excellent property of upper bounding the relationship between input perturbation and output variation for a given distance (generally L2 norm), thus it is considered as a metric to evaluate the robustness of neural networks to small perturbations \cite{luxburg2004distance, virmaux2018lipschitz, bartlett2017spectrally}. However, computing the exact Lipschitz constant of neural networks in the knowledge distillation process is a NP-hard problem \cite{virmaux2018lipschitz}. To solve this problem, we propose a feasible and effective method to approximate the Lipschitz constants in KD.

We first define the affine function for the $k$-th layer $T^k:\mathbf{fm}^{k-1} \longmapsto \mathbf{fm}^{k}$, in which $\mathbf{fm}^{k-1} \in \mathbb{R}^{d_{k-1}}$ and $\mathbf{fm}^{k} \in \mathbb{R}^{d_{k}}$ are the feature maps out of the $(k-1)$th and the $k$th layer, respectively.

By \textbf{Lemma} 1 as in Supplemetary Appendix, we have $\Vert T^k\Vert_{Lip}= {\sup}_{\mathbf{fm}} \Vert \nabla T^k(\mathbf{fm}) \Vert_{SN}$, where $ \Vert \cdot \Vert_{SN}$ is the spectral norm of matrix. And the matrix spectral norm $ \Vert \cdot \Vert_{SN}$ is formally defined by
\begin{equation}
    \Vert \mathbf{W} \Vert_{SN} \triangleq \max \limits_{\mathbf{x}:\mathbf{x}\neq \mathbf{0}} \frac{\Vert \mathbf{W} \mathbf{x} \Vert_2}{\Vert \mathbf{x} \Vert_2} = \max \limits_{\Vert\mathbf{x}\Vert_2 \leq  1}{\Vert \mathbf{W} \mathbf{x} \Vert_2},
\end{equation}
where the spectral norm of matrix $\mathbf{W}$ is equivalent to its largest singular value. Thus, for the linear layer $T^k (\mathbf{fm}^{k-1}) = \mathbf{fm}^{k}$, based on Lemma 2 in Supplemetary Appendix, its Lipschitz constant is given by
\begin{equation}
    \Vert T^k\Vert_{Lip} = {\sup}_{\mathbf{fm}} \Vert \nabla T^k(\mathbf{fm}) \Vert_{SN} = \Vert \mathbf{W} \Vert_{SN}.
\end{equation}
Additionally, most activation functions such as ReLU, Leaky ReLU, Tanh, Sigmoid as well as max-pooling, have a Lipschitz constant equal to 1. As for other common neural network layers such as dropout, batch normalization and other pooling methods, they all have simple and explicit Lipschitz constants \cite{goodfellow2016deep}. This fixed Lipschitz constant property renders our derivation applicable to most network architectures, such as ResNet \cite{he2016deep} and MobileNet \cite{howard2017mobilenets}.

Thereafter, we use the inequality (concluded by Eq. 7 in \cite{bartlett2017spectrally}) $\Vert T^k \circ T^{k+1} \Vert_{Lip} \leq \Vert T^k \Vert_{Lip}\cdot \Vert T^{k+1} \Vert_{Lip}$ to derive the following bound for $\Vert f \Vert_{Lip}$:
\begin{equation}
\begin{split}
    \!\!\!\!\Vert f \Vert_{Lip} &\leq  \Vert T^L \Vert_{Lip} \cdot \Vert \sigma \Vert_{Lip} \cdot \Vert T^{L-1} \Vert_{Lip} \cdots \cdot \Vert T^1 \Vert_{Lip}\!\!\!\! \\
    &= \prod_{k=1}^{L}\Vert T^k \Vert_{Lip} = \prod_{k=1}^{L}\Vert \mathbf{W}^{k}\Vert_{SN}.
\end{split}
\label{eq:6}
\end{equation}

In this way, we transfer the teacher's Lipschitz constant to the student through a sequence of spectral norm of intermediate layers in the network. Moreover, the upper bound of Lipschitz constant also ensures the quality of knowledge to be transferred.

\subsection{Transmitting Matrix}
\label{sec:33}
Given the derived tight upper bound of the Lipschitz constant, we design a novel loss to distill Lipschitz continuity from teacher to student by narrowing the distance between corresponding $\Vert \mathbf{W}_T^{k}\Vert_{SN}$ and $\Vert \mathbf{W}_S^{k}\Vert_{SN}$ down. The first problem is how to calculate each spectral norm. Calculating the spectral norm of weight matrix $\mathbf{W}^k$ in neural networks by SVD is inaccessible. Specifically, for the complex network structures such as convolutions layers or residual modules, though they can be re-arranged matrix-wisely, their spectral norm's computation is impractical. Therefore, we propose using Transmitting Matrix (TM) to bypass the complicated calculation of the spectral norm $\mathbf{W}^k$. This approximate calcuation allows feasible computation to distill Lipschitz constant and its further use as a loss function.

For training data of batch size $N$, after a forward process for the $(k$-1)th layer, we have a batch of corresponding feature maps as
\begin{equation}
    \!\!\!\!\mathbf{FM}^{k-1} = (\mathbf{fm}^{k-1}_1,\mathbf{fm}^{k-1}_2,\cdots,\mathbf{fm}^{k-1}_n) \in \mathbb{R}^{{d_{k-1}}\times N},
\end{equation}
where $\mathbf{W}^k \mathbf{FM}^{k-1}= \mathbf{FM}^{k}$ for each $k \in \{1,\dots,L\}$.

Studies~\cite{chen2017exemplar, tung2019similarity} about similarity of feature maps illustrate that for well-trained networks, their batch of feature maps in the same layer $\{\mathbf{fm}^{k-1}_i\}, i \in \{1,\dots,n\}$ have strong mutual linear independence. We formalize the relevance of feature maps in the same layer as
\begin{align}
    \forall i\neq j\in\{1,\cdots,N\}, (\mathbf{fm}^{k-1}_i)^{\mathsf{T}} \mathbf{fm}^{k-1}_j \approx 0, \\
    \forall i \in\{1,\cdots,N\}, (\mathbf{fm}^{k-1}_i)^{\mathsf{T}} \mathbf{fm}^{k-1}_i \neq 0.
\label{eq:9}
\end{align}

We further normalize the feature maps by $\forall i \in\{1,\cdots,N\}, \mathbf{fm}^{k-1}_i = \frac{\mathbf{fm}^{k-1}_i}{\Vert (\mathbf{fm}^{k-1}_i)\Vert_2}$ such that a batch of feature maps can be expressed in a vector representation that
\begin{equation}
\label{eq:indenpendency}
    (\mathbf{FM}^{k-1})^{\mathsf{T}}\mathbf{FM}^{k-1} \approx \mathbf{I},
\end{equation}
where $\mathbf{I}$ is an unit matrix. 

With all the aforementioned equations, we are ready to define the transmitting matrix $\mathbf{TM}^k$ for calculating the spectral norm of matrix $\mathbf{W}^k$ as calculating the spectral norm of $\mathbf{TM}^k \triangleq \left[(\mathbf{FM}^{k-1})^{\mathsf{T}} \mathbf{FM}^{k}\right]^{\mathsf{T}}\!\! \left[(\mathbf{FM}^{k-1})^{\mathsf{T}} \mathbf{FM}^{k}\right]$

%Having all the equations, we are ready to define the transmitting matrix $\mathbf{TM}^k$ for calculating the spectral norm of matrix $\mathbf{W}^k$:
\begin{equation}
\label{eq:tm}
\begin{split}
    % & \\
    &=  (\mathbf{W}^k \mathbf{FM}^{k-1})^{\mathsf{T}} (\mathbf{FM}^{k-1})
     \left[(\mathbf{FM}^{k-1})^{\mathsf{T}}  \mathbf{W}^k \mathbf{FM}^{k-1}\right]\\
    &=  (\mathbf{FM}^{k-1})^{\mathsf{T}} (\mathbf{W}^k)^{\mathsf{T}} (\mathbf{FM}^{k-1})
     (\mathbf{FM}^{k-1})^{\mathsf{T}}  \mathbf{W}^k \mathbf{FM}^{k-1}.\\
\end{split}
\end{equation}
Eq.~\ref{eq:indenpendency} and \ref{eq:tm} together yield the result as
\begin{equation}
\label{eq:main}
     \mathbf{TM}^k \approx (\mathbf{FM}^{k-1})^{\mathsf{T}} ({\mathbf{W}^k}^{\mathsf{T}}\mathbf{W}^k) \mathbf{FM}^{k-1}.
\end{equation}
\noindent\textbf{Theorem 1.} If matrix $\mathbf{U}$ is an orthogonal matrix, such that $\mathbf{U}^\mathsf{T}\mathbf{U} = \mathbf{I}$, where $\mathbf{I}$ is an unit matrix, the largest eigenvalues of $\mathbf{U}^\mathsf{T} \mathbf{H} \mathbf{U}$ and $\mathbf{H}$ are equivalent.
\begin{equation}
    \sigma_1( \mathbf{U}^\mathsf{T} \mathbf{H} \mathbf{U}) = \sigma_1( \mathbf{H}),
\end{equation}
where $\sigma_1(\cdot)$ is the largest eigenvalue of a matrix.
Based on Theorem 1 and Eq. \ref{eq:main}, our defined transmitting matrix $\mathbf{TM}^k$ has the same largest eigenvalue with ${\mathbf{W}^k}^{\mathsf{T}}\mathbf{W}^k$, \ie $\sigma_1(\mathbf{TM}^k) = \sigma_1({\mathbf{W}^k}^{\mathsf{T}}\mathbf{W}^k)$ . Thus, combining the definition of spectral norm $\Vert \mathbf{W}^k\Vert =  \sigma_1({\mathbf{W}^k}^{\mathsf{T}}\mathbf{W}^k)$, we can achieve the spectral norm of matrix $\mathbf{W}^k$ by calculating the largest eigenvalue of $\mathbf{TM}^k$, $\sigma_1(\mathbf{TM}^k)$, which is solvable.

For networks with more complicated layers such as residual blocks, by considering the block as an affine mapping from front feature maps to back feature maps, this approximation is applicable to calculate the spectral norm block-by-block instead of layer-by-layer, which makes our spectral norm calculation more efficient. To this end, we define the Transmitting Matrix $\mathbf{TM}$ for residual blocks as
\begin{equation}
     \mathbf{TM}_m^k \triangleq \left[(\mathbf{FM}^{f})^{\mathsf{T}} \mathbf{FM}^{l}\right]^{\mathsf{T}} \left[(\mathbf{FM}^{f})^{\mathsf{T}} \mathbf{FM}^{l}\right], \\
\end{equation}
where the $\mathbf{FM}^{f}$ and $\mathbf{FM}^{l}$ are the front feature maps and latter feature maps of a residual block.

\subsection{Approximating the Spectral Norm with Power Iteration Method}
\label{sec:34}
Following the aforementioned steps, we next need to calculate the spectral norms of two matrices (teacher and student) and then calculate the loss between those two. The intuitive approach is using SVD to compute the spectral norm, which results in overloaded computation. Importantly, the SVD calculation is non-differentiable, making it impossible to train the deep networks. Instead of using SVD, we utilize power iteration method \cite{golub2000eigenvalue, yoshida2017spectral, miyato2018spectral} to approximate the spectral norm of the targeted matrix with a small trade-off of accuracy, as presented in Algorithm \ref{alg:pi}. 
\begin{algorithm}  
	\caption{Compute spectral norm using power iteration}  
	\begin{algorithmic}[1] 
		\Require targeted matrix $\mathbf{TM}$, stop condition $res_{stop}$.
		\Ensure the spectral norm of matrix $\mathbf{TM}$, $\Vert \mathbf{TM} \Vert_{SN}$.
		\State initialize $\mathbf{v}_0 \in \mathbb{R}^m$ with a random vector.
	    \While{$res\geq res_{stop}$}
	    	\State $\mathbf{v}_{i+1} \gets \mathbf{TM}\mathbf{v}_{i} \bigl / \Vert \mathbf{TM}\mathbf{v}_{i}\Vert_2$
	    	\State $res = \Vert \mathbf{v}_{i+1} - \mathbf{v}_{i}\Vert_2$
		\EndWhile
		\State \Return{$\Vert \mathbf{TM} \Vert_{SN} = \mathbf{v}_{i+1}^{\mathsf{T}} \mathbf{TM} \mathbf{v}_{i} $}
	\end{algorithmic}  
	\label{alg:pi}
\end{algorithm}

In this way, we have a feasible approach to calculate the spectral norms of $\mathbf{TM}$s which can faithfully approximate the Lipschitz constant of networks.
% \vspace{-0.3in}

\subsection{Overall Loss Function}

By using Algorithm \ref{alg:pi}, we obtain the spectral norms for teacher and student networks, respectively: $\Vert \mathbf{TM}_T^i \Vert_{SN}$ and $\Vert \mathbf{TM}_S^i \Vert_{SN}$ for each $i\in \{1, \dots, L\}$. We define our novel lipschitz continuity loss function $\mathcal{L}_{Lip}$ as
\begin{equation}
    \mathcal{L}_{Lip} = \sum_{i=1}^{L-1}(\frac{\Vert \mathbf{TM}_T^i \Vert_{SN} - \Vert \mathbf{TM}_S^i \Vert_{SN}}{\beta^{L-1-i}})^2,
    \label{equ:15}
\end{equation}

where $\beta$ is a coefficient greater than $1$. Hence, the ${\beta^{L-1-i}}$ decreases with $i$ increasing and consequently the $\frac{\Vert \mathbf{TM}_T^i \Vert_{SN} - \Vert \mathbf{TM}_S^i \Vert_{SN}}{\beta^{L-1-i}}$ increases. In this way, we give more weight on higher layer features since they are closer to the features performing tasks.

Combined with the cross entropy loss $\mathcal{L}_{CE}$ and vanilla knowledge distillation loss $\mathcal{L}_{KD}$, we are ready to propose our novel loss function as
\begin{equation}
    \mathcal{L} = \frac{\lambda}{2}\cdot\mathcal{L}_{Lip} + \mathcal{L}_{KD} + \mathcal{L}_{CE},
    \label{equ:16}
\end{equation}
where $\lambda$ is used to control the degree of distilling the Lipschitz constant. We use $\frac{\lambda}{2}$ because when taking derivative of  $\mathcal{L}_{Lip}$, the denominator part can be easily eliminated.

\subsection{Explaining the Effectiveness from a Regularization Perspective}
\label{sec:overfitting}
The derivative of the loss function $\mathcal{L}$ with respect to $W$:
\begin{equation}
\begin{split}
    &\frac{\partial\mathcal{L}}{\partial \mathbf{W}} = \frac{\partial (\mathcal{L}_{CE} + \mathcal{L}_{KD})}{\partial \mathbf{W}} + \frac{\partial (\mathcal{L}_{Lip})}{\partial \mathbf{W}}\\
    &= \mathbf{M} - \lambda\sum_{i=1}^{L-1}(\frac{\Vert \mathbf{TM}_T^i \Vert_{SN} - \Vert \mathbf{TM}_S^i \Vert_{SN}}{\beta^{L-1-i}})\frac{\partial \Vert \mathbf{TM}^i_S\Vert_{SN}}{\partial \mathbf{W}} \\
    &\approx \mathbf{M} - \lambda\sum_{i=1}^{L-1}(\frac{\Vert \mathbf{TM}_T^i \Vert_{SN} - \Vert \mathbf{TM}_S^i \Vert_{SN}}{\beta^{L-1-i}})\frac{\partial \Vert W^i_S\Vert_{SN}}{\partial \mathbf{W}} \\
    &= \mathbf{M} - \lambda\sum_{i=1}^{L-1}(\frac{\Vert \mathbf{TM}_T^i \Vert_{SN} - \Vert \mathbf{TM}_S^i \Vert_{SN}}{\beta^{L-1-i}}) \mathbf{u}_1^i (\mathbf{v}_1^i)^{\mathsf{T}},
\end{split}
\label{eq:18}
\end{equation}
where $\mathbf{M} \triangleq \frac{\partial (\mathcal{L}_{CE} + \mathcal{L}_{KD})}{\partial \mathbf{W}}$, $ \mathbf{u}_1^i$ and $ \mathbf{v}_1^i$ are respectively the first left and right singular vectors of $\mathbf{W}_S^i$ . For $\mathbf{W}_S^i$, using SVD, we have
\begin{equation}
    \mathbf{W}_S^i = \sum_{j=1}^{d_i} \sigma_j(\mathbf{W}_S^i) \mathbf{u}_j^i \mathbf{v}_j^i,
    \label{eq:19}
\end{equation}
where $d_i$ is the rank of $ \mathbf{W}_S^i$, $\sigma_j(\mathbf{W}_S^i)$ is the $j$-th biggest singular value, $\mathbf{u}_j^i$ and $\mathbf{v}_j^i$ are correspondingly left and singular vectors, respectively. 

In Eq. \ref{eq:18}, the first term $\mathbf{M}$ is the same as the derivative of the loss function of vanilla knowledge distillation. As for the second term, based on Eq. \ref{eq:19}, it can be seen as the regularization term penalizing the vanilla knowledge distillation loss with an adaptive regularization coefficient 
\begin{equation}
   \gamma \triangleq \lambda\frac{\Vert \mathbf{TM}_T^i \Vert_{SN} - \Vert \mathbf{TM}_S^i \Vert_{SN}}{\beta^{L-1-i}}, 
\end{equation}
which constrains the weights of the student networks by utilizing teacher networks' $\Vert \mathbf{TM}_T^i \Vert_{SN}$ as a prior supervision information. In other words, our method prevents the student networks from trapping into local minima. In this way, it ensures better training of student networks. We demonstrate performance by designing a corresponding experiment in Section \ref{exp:mitigate}, showing that our proposed method prevents student networks from over-fitting the dataset.

\section{Experiments}
In this section, we conducted experiments on three computer vision tasks, image classification, object detection, and segmentation, to validate the effectiveness of our proposed distillation method. In addition to comparing our method with the state-of-the-art methods, we also designed a series of ablation studies to verify the effectiveness and highlight the regularization property of our proposed technique. All experiments are implemented using PyTorch \cite{paszke2019pytorch}.

\begin{table*}[!t]\small
\centering
\scalebox{1.07}{
\begin{tabular}{c|c|cc|ccc}
    \multirow{2}{*}{Setup} & \multirow{2}{*}{Compression type}  & \multirow{2}{*}{Teacher network}  & \multirow{2}{*}{Student network}     & \# of params& \# of params & Compress \\ 
    &&&&teacher&student&ratio\\
    \specialrule{0.8pt}{0pt}{0pt}
    (a)   & Depth                  & WideResNet 28-4      & WideResNet 16-4     & 5.87M                                                          & 2.77M                                                       & 47.2\%                                                   \\
    (b)   & Channel                & WideResNet 28-4      & WideResNet 28-2     & 5.87M                                                          & 1.47M                                                       & 25.0\%                                                   \\
    (c)   & Depth \& channel       & WideResNet 28-4      & WideResNet 16-2     & 5.87M                                                          & 0.70M                                                       & 11.9\%                                                   \\
    (d)   & Different architecture & WideResNet 28-4      & ResNet 56           & 5.87M                                                          & 0.86M                                                       & 14.7\%                                                   \\
    (e)   & Different architecture & PyramidNet-200 (240) & WideResNet 28-4     & 26.84M                                                         & 5.87M                                                       & 21.9\%                                                   \\
    (f)   & Different architecture & PyramidNet-200 (240) & PyramidNet-110 (84) & 26.84M                                                         & 3.91M                                                       & 14.6\%                                                   \\
    (g)   & Different architecture & PyramidNet-200 (240) & ResNet 56           & 26.84M                                                         & 0.86M                                                       & 5.8\%                                                    \\ 
\end{tabular}
}
\caption{Seven experimental settings with different network topological structures on CIFAR-100.}
\label{tabel:setting}
\end{table*}

\begin{table*}[!t]\footnotesize
\centering
\scalebox{1.22}{
\begin{tabular}{c|c|ccccccccc|c}
    \multirow{2}{*}{Setup} & \multirow{2}{*}{Teacher} & \multirow{2}{*}{Baseline} & KD   & FitNets  & AT    & Jacobian  & FT    & AB    & OFD    & AFD   & LONDON         \\ 
    &&& \cite{hinton2015distilling} & \cite{romero2014fitnets} & \cite{zagoruyko2016paying} & \cite{srinivas2018knowledge} & \cite{kim2018paraphrasing} & \cite{heo2019knowledge} & \cite{heo2019comprehensive} & \cite{wang2019pay} & (\textbf{ours})\\
    \specialrule{0.8pt}{0pt}{0pt}
    (a)   & 21.09   & 22.72    & 21.69 & 21.85   & 22.07 & 22.18    & 21.72 & 21.36 & 20.89 & 21.15 & \textbf{20.33} \\
    (b)   & 21.09   & 24.88    & 23.43 & 23.94   & 23.80 & 23.70    & 23.41 & 23.19 & 21.98 & 21.79 & \textbf{20.71} \\
    (c)   & 21.09   & 27.32    & 26.47 & 26.30   & 26.56 & 26.71    & 25.91 & 26.02 & 24.08 & 24.21 & \textbf{23.46} \\
    (d)   & 21.09   & 27.68    & 26.76 & 26.35   & 26.66 & 26.60    & 26.20 & 26.04 & 24.44 & 24.67 & \textbf{23.78} \\
    (e)   & 15.57   & 21.09    & 20.97 & 22.16   & 19.28 & 20.59    & 19.04 & 20.46 & 17.80 & 18.24 & \textbf{17.54} \\
    (f)   & 15.57   & 22.58    & 21.68 & 23.79   & 19.93 & 23.49    & 19.53 & 20.89 & 18.89 & 19.32 & \textbf{18.21} \\
    (g)   & 15.57   & 27.68    & 26.82 & 26.10   & 26.64 & 26.43    & 26.29 & 25.70 & 24.49 & 24.53 & \textbf{23.52} \\ 
\end{tabular}
}
\caption{Top-1 error rates (\%) of 7 different combinations in Table~\ref{tabel:setting} on CIFAR-100 test set. Lower is better. \textit{Baseline} represents a result without distillation. For all the results we used author-provided code or author-reported results. Each result is averaged over 5 runs.}
\label{tabel:cifar}
% \vspace{-0.1in}
\end{table*}

\subsection{Classification}
We chose CIFAR-100 \cite{krizhevsky2009learning} for classification. This is because it is commonly used for comparing KD methods and its relatively small size provides flexibility of implementing different combinations of teacher and student architectures. Besides CIFAR-100, we conducted experiments on ImageNet \cite{deng2009imagenet}, a larger dataset, to verify the stability of our distillation method.

\noindent\textbf{CIFAR-100} \cite{krizhevsky2009learning} is the most widely-used image classification dataset, which consists of 50K training images and 10K testing images of size $32\times 32$ divided into 100 classes. Specifically, we designed various combinations of architectures for teacher and student networks. Table~\ref{tabel:setting} summarizes the settings of each experiment, model size and compression ratio, involving architectures such as Residual Network (ResNet) \cite{he2016deep}, Wide Residual Network (WideResNet) \cite{zagoruyko2016wide}, and Deep Pyramidal Residual Networks (PyramidNet) \cite{han2017deep}. Experimental results of different settings are shown in Table \ref{tabel:cifar}, where it is obvious that our method achieves state-of-the-art performance in all seven settings, for both depth and channel compression (\textbf{a, b, c}) and different architectures (\textbf{d, e, f, g}). Especially, in the setting of depth compression and channel compression (a) and (b), the student networks trained by LONDON even outperform the teacher networks, which further demonstrates the efficacy of our Lipschitz continuity method as a regularization function.

Overall, our proposed method consistently shows comparable or better performance regardless of different compression rates or other network architecture types, which endows our approach with more implementation flexibility. We noted exciting improvements in student networks along with a high compression ratio. Therefore, our results present the potential of using Lipschitz continuity distillation to compress large networks into more resource-efficient ones with acceptable accuracy drop. For example, when the setting (g) is a $17 \times$ compression from teacher network to student network with completely different architecture, the student network still benefits from the teacher network via our method. 
% Note that the error rate of 17.54\% in (e) is better than any network reported in the paper of Wide Residual Network \cite{zagoruyko2016wide}. 
In general, our proposed method can be applied to small networks (fewer parameters) and large networks with satisfactory performance.

\noindent\textbf{ImageNet} \cite{deng2009imagenet} is a large-scale dataset with 1.2 million training images and 50k validation images divided into 1,000 classes. Compared to other classification datasets such as CIFAR-100, ImageNet has greater diversity, and its image is larger in scale (average $469 \times 387$). For all experiments, we reported both the top-1 and top-5 accuracies. Images are cropped to the size of $224 \times 224$ for training and validation. The student networks are trained for 100 epochs, and the learning rate begins at 0.1 multiplied by 0.1 at every 30 epochs. To ensure a fair comparison, we used the pre-trained models in the PyTorch library as the teacher networks. Two combinations of network architectures are settled for demonstration. For the first combination, we chose ResNet152 \cite{he2016deep} as the teacher network and ResNet50 as the student network. As the second one, for testing the knowledge distillation capacity across different network architectures, we chose ResNet50 as the teacher network, and MobileNet \cite{howard2017mobilenets} as the student network. The results are displayed in Table \ref{table:imagenet}. Compared to strong methods such as \cite{heo2019knowledge,heo2019comprehensive}, our method still exhibits a great improvement. In particular, our method makes ResNet50 outperform the teacher network ResNet152, which is a remarkable achievement. Besides, regarding the compression ability, our method makes a considerable improvement in the lightweight architecture, MobileNet, where the error rate of 27.64\% of our method is better than any network reported in the paper of MobileNet \cite{howard2017mobilenets}.

\begin{table}[!ht]\small
\centering
\resizebox{0.50\textwidth}{!}{%
\begin{tabular}{p{1.5cm}c|p{2.5cm}|cc}
    \multirow{2}{*}{Network}   & \# of params & \multirow{2}{*}{Method}   & Top-1          & Top-5\\ 
    &(ratio)&&error&error\\
    \specialrule{0.8pt}{0pt}{0pt}
    ResNet152 & 60.19M  & Teacher  & 21.69          & 5.95          \\ 
    \specialrule{0.8pt}{0pt}{0pt}
      &                                                                                    & Baseline & 23.72          & 6.97          \\
      &                                                                                    & KD \cite{hinton2015distilling}      & 22.85          & 6.55          \\
      &                                                                                    & AT  \cite{zagoruyko2016paying}     & 22.75          & 6.35          \\
    \multirow{2}{*}{ResNet50}  & 25.56M                                                                             & FT \cite{kim2018paraphrasing}      & 22.80          & 6.49          \\
      & (42.5\%)                                                                           & AB  \cite{heo2019knowledge}    & 23.47          & 6.94          \\
      &                                                                                    & OFD  \cite{heo2019comprehensive}    & 21.65          & 5.83          \\
      &                                                                                    & AFD  \cite{wang2019pay}    & 22.08          & 6.30          \\
      &                                                                                    & LONDON (\textbf{ours})   & \textbf{21.12} & \textbf{5.47} \\ 
    \specialrule{0.8pt}{0pt}{0pt}
    ResNet50  & 25.56M   & Teacher  & 23.84          & 7.14          \\
    \specialrule{0.8pt}{0pt}{0pt}
      &                                                                                    & Baseline & 31.13          & 11.24         \\
      &                                                                                    & KD \cite{hinton2015distilling}      & 31.42          & 11.02         \\
      &                                                                                    & AT \cite{zagoruyko2016paying}      & 30.44          & 10.67         \\
    \multirow{2}{*}{MobileNet} & 4.23M                                                                              & FT  \cite{kim2018paraphrasing}     & 30.12          & 10.50         \\
      & (16.5\%)                                                                             & AB  \cite{heo2019knowledge}     & 31.11          & 11.29         \\
      &                                                                                    & OFD \cite{heo2019comprehensive}     & 28.75          & 9.66          \\
      &                                                                                    & AFD \cite{wang2019pay}     & 28.61          & 9.81          \\
      &                                                                                    & LONDON (\textbf{ours})  & \textbf{27.64} & \textbf{8.97} \\ 
\end{tabular}
}
\caption{Top-1 and Top-5 error rates (in percentage) of different combinations of the student and teacher's network structures on ImageNet validation set. 'baseline' represents a result without distillation technique. Lower is better.}
\label{table:imagenet}
% \vspace{-0.2in}
\end{table}

\begin{table}[!t]\small
\centering
\scalebox{1.02}{
\begin{tabular}{p{2.1cm}c|p{2.3cm}|c}
\multirow{2}{*}{Network}   & \# of params & \multirow{2}{*}{Method}   & \multirow{2}{*}{mAP}\\ 
&(ratio)&&\\
\specialrule{0.8pt}{0pt}{0pt}
ResNet50-SSD  & 36.7M        & Teacher  & 76.79                           \\ 
\specialrule{0.8pt}{0pt}{0pt}
              &              & Baseline & 71.61                           \\
\multirow{2}{*}{ResNet18-SSD}& 20.0M        & OFD  \cite{heo2019comprehensive}    & 73.08                            \\
              & (54.5\%)              & AFD  \cite{wang2019pay}    & 72.78               \\
              &              & LONDON (\textbf{ours})  & \textbf{73.82} \\ \hline
              &              & Baseline & 67.58                           \\
\multirow{2}{*}{MobileNet-SSD} & 6.5M         & OFD  \cite{heo2019comprehensive}    & 68.54                           \\
              &(18.7\%)              & AFD  \cite{wang2019pay}   & 68.63                           \\
              &              & LONDON (\textbf{ours})  & \textbf{69.09} \\ 
\end{tabular}}
\caption{Object detection results in PASCAL VOC2007 testing set. Results are described in mean Average Precision (mAP). Higher is better.}
\label{table:detection}
% \vspace{-1.2em}
\end{table}

\subsection{Object Detection}
We applied our proposed method on the most popular high-speed detector, Single Shot Detector (SSD) \cite{liu2016ssd}. All models are trained with the training set of VOC2007 and VOC2012 \cite{everingham2015pascal} where the backbone networks are pre-trained using the ImageNet dataset. All models are trained for 120k iterations with a batch size of 32. We set the SSD trained with no distillation as our baseline and SSD detector with ResNet50 as the teacher network. As for the student networks, we used SSD with ResNet18, or MobileNet \cite{howard2017mobilenets}. We evaluated the detection performance in the VOC2007 testing set. The result is presented in Table \ref{table:detection}. Both trained student networks outperform other methods. This implies that our method can be applied to object detector. Furthermore, we found that the distillation between similar structures has better quality than the different ones by comparing the performance of ResNet18 to MobileNet.

\begin{table}[!t]\small
\centering
\scalebox{1.0}{%
\begin{tabular}{p{2cm}c|p{2.5cm}|c}
\multirow{2}{*}{Backbone}   & \# of params & \multirow{2}{*}{Method}   & \multirow{2}{*}{mIoU}\\ 
&(ratio)&&\\
\specialrule{0.8pt}{0pt}{0pt}
ResNet101 & 59.3M        & Teacher  & 77.39          \\
\specialrule{0.8pt}{0pt}{0pt}
          &              & Baseline  & 71.79          \\
ResNet18  & 16.6M        & OFD \cite{heo2019comprehensive}     & 73.24          \\
          & (28.0\%)             & AFD \cite{wang2019pay}    & 72.81          \\
          &              & LONDON (\textbf{ours})  & \textbf{73.62} \\ \hline
          &              & Baseline  & 68.44          \\
MobileNet & 5.8M         & OFD  \cite{heo2019comprehensive}    & 71.36          \\
          & (9.8\%)             & AFD  \cite{wang2019pay}    & 71.56          \\
          &              & LONDON (\textbf{ours})   & \textbf{71.97} \\ 
\end{tabular}}
\caption{Semantic segmentation on the PASCAL VOC 2012 testing set. Results are described in mean Intersection over Union (mIoU). Higher is better.}
\label{table:segmentation}
% \vspace{-0.2in}
\end{table}
\subsection{Semantic Segmentation}
In this section, we conducted knowledge distillation on semantic segmentation task. It is worth noting that implementing KD on semantic segmentation is extremely difficult for the penultimate feature maps of the segmentation model, which has higher dimensions than common network architectures. In particular, the widely-used DeepLabV3+ \cite{ChenPSA17} is taken as our study case for semantic segmentation. We used DeepLabV3+ with the backbone of ResNet101 as the teacher, and DeepLabV3+ based on ResNet18\cite{he2016deep} and MobileNetV2 \cite{howard2017mobilenets} as the students. The results shown in Table \ref{table:segmentation} provide clear evidence that our proposed method can greatly improve the performance of both ResNet18 and MobileNet.

In general, most KD studies are only experimentally justified over the task of image classification. In our case, experiments on detection and segmentation verify that our method can be applied to not only image classification but also other computer vision tasks. The flexibility without significant model modifications is an advantage of our high-level knowledge distillation so that our proposed method has a wide range of potential applications.

\subsection{Analyses}

\label{exp:mitigate}
\begin{figure}[!t]
    \centering
    \includegraphics[width=0.48\textwidth]{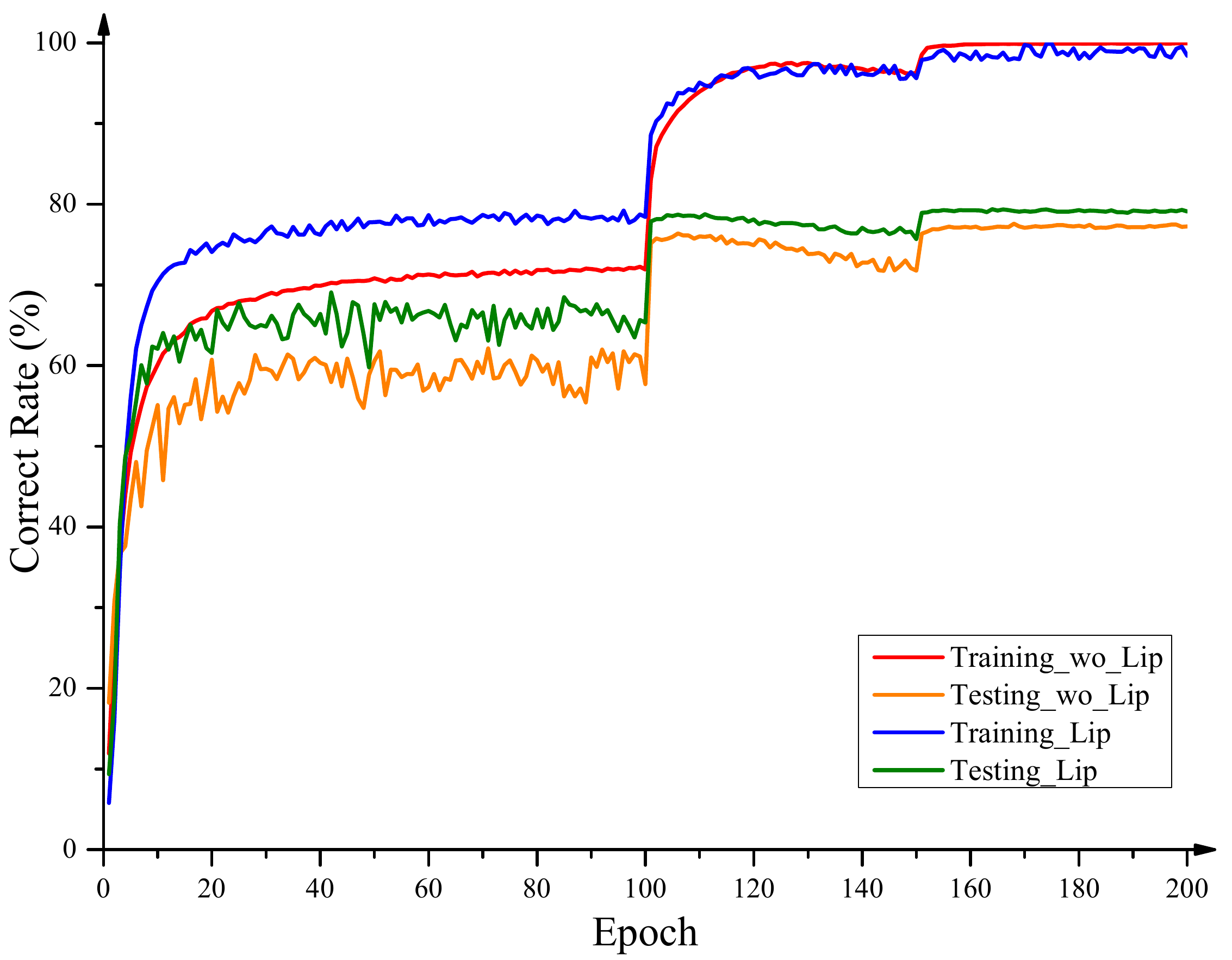}
    \caption{Our proposed loss can mitigate overfitting.}
    \label{fig:mitigating}
    % \vspace{-0.2in}
\end{figure}

\begin{table}[!t]\small
\scalebox{0.97}{
\begin{tabular}{c|cccccc}
\diagbox{Pair}{$\lambda$} & 0     & 0.1   & 0.4   & 1.6            & 3.2            & 6.4   \\ \specialrule{0.8pt}{0pt}{0pt}
(a)                                                 & 21.69 & 21.36 & 21.54 & 21.11          & \textbf{20.33} & 21.87 \\
(b)                                                 & 23.43 & 22.04 & 22.05 & 21.88          & \textbf{21.48} & 22.35 \\
(c)                                                 & 26.47 & 24.39 & 23.77 & \textbf{23.56} & 23.62          & 24.87 \\
(d)                                                 & 27.68 & 24.18 & 24.42 & 23.82          & \textbf{23.78} & 25.22 \\ 
\end{tabular}}
\caption{Ablation study of our method. The results are presented in the form of error rate (\%). Lower is better.}
\label{table:ablation}
% \vspace{-1.5em}
\end{table}

\noindent \textbf{Mitigate Overfitting.} As demonstrated in Section \ref{sec:overfitting}, our Lipschitz distillation loss can be seen as a regularization term, which constrains the search space around the point inferred by the teacher so as to prevents overfitting the target dataset. To consolidate this theoretical demonstration, we design a corresponding experiment. We use the setting (b) in Table \ref{tabel:setting} to study this regularization phenomenon. The results are shown in Figure~\ref{fig:mitigating}. 
% \syz{Even CIFAR-100 is a simple dataset, which is easily overfitted (models reaching near 100\% on training set). } 
It is noteworthy that when turning off the Lipschitz continuity loss module, the performance on the validation set drops while the training correct rate stays at the same level. This overfit-reduction phenomenon verifies that our proposed method improves the student network training by regularization.
%\vspace{-5pt}

\noindent \textbf{Ablative Experiments.} We conducted an ablation study of our proposed method in CIFAR-100 with the teacher and student architecture pairs in Table \ref{tabel:setting}. By adjusting the coefficient $\lambda$ in the loss function $\mathcal{L}_{London}$ (Eq. \ref{equ:15}, \ref{equ:16}), where $\lambda = 0$ equals to no Lipschitz continuity distilled as our baseline. The results are shown in Table \ref{table:ablation}. With $\lambda$ increasing, the performance improvements show the effectiveness of our designed Lipschitz continuity loss. However, when the ratio of $\mathcal{L}_{Lip}$ in $\mathcal{L}_{London}$ is greater than 20\% (on average), LONDON's performance drops. A well-trained student network should have both the ability to align low-level feature maps and capture the high-level information. Therefore, we believe that putting too much weight on high-level and universal information loses the aligning ability that the network would have.
% \vspace{-8pt}

\section{Conclusion}
We investigate the knowledge distillation and Lipschitz continuity of neural networks. Specifically, we present a novel KD method, named LONDON, which numerically calculates and transfers the Lipschitz constant as knowledge. Compared to standard KD methods considering neural networks as black-boxes, our KD method captures the functional property of neural networks as high-level knowledge for training student networks, which further prevents the students networks from overfitting the datasets by extending the representational capability of KD.  %The experimental results demonstrate that our method achieves the state-of-the-art performance over the tasks of classification, object detection and segmentation. 

\noindent \textbf{Acknowledgements.} This research was partially supported by NSF CNS-1908658, NeTS-2109982 and the gift donation from Cisco. This article solely reflects the opinions and conclusions of its authors and not the funding agents.
% In future, we plan to continue exploring Lipschitz continuity KD. Since this high-level knowledge can be seen as a regularization term, it makes our method model-agnostic. Therefore the module of distilling Lipschitz continuity can be flexibly implemented on other network compression paradigms by simply piling up our designed module on the existing state-of-the-art architectures.

%-------------------------------------------------------------------------------------
{\small
\bibliographystyle{ieee_fullname}
\bibliography{manuscript}
}

\clearpage
\section{Appendix}
\subsection{Lemma 1.}
\label{lemma:1}
Based on Rademacher's theorem \cite{federer2014geometric}, for the functions restricted to some neighborhood around any point is Lipschitz, their Lipschitz constant can be calculated by their differential operator.

\noindent\textbf{Lemma 1.} If a function $f : \mathbb{R}^{n} \longmapsto \mathbb{R}^{m}$ is a locally Lipschitz continuous function, then $f$ is differentiable almost everywhere. Moreover, if $f$ is Lipschitz continuous, then
\begin{equation}
    \Vert f\Vert_{Lip} = \sup_{\mathbf{x}\in\mathbb{R}^{n}}\Vert  \nabla_{\mathbf{x}} f \Vert_2
\end{equation}
where $\Vert \cdot\Vert_2$ is the L2 matrix norm.

\subsection{Lemma 2.}
\label{lemma:2}
\noindent\textbf{Lemma 2.} 
Let $\mathbf{W} \in \mathbb{R}^{m \times n}, \mathbf{b} \in \mathbb{R}^{m}$ and $T(\mathbf{x}) = \mathbf{W}\mathbf{x} + \mathbf{b}$ be an linear function. Then for all $\mathbf{x} \in \mathbb{R}^{n}$, we have
\begin{equation}
    \nabla g(\mathbf{x}) = \mathbf{W}^{\mathsf{T}}\mathbf{W}\mathbf{x}
\end{equation}
where $g(\mathbf{x}) = \frac{1}{2}\Vert f(\mathbf{x}) - f(\mathbf{0})\Vert_2^2$.

\noindent\textbf{Proof.} By definition, $g(\mathbf{x}) = \frac{1}{2}\Vert f(\mathbf{x}) - f(\mathbf{0})\Vert_2^2 = \frac{1}{2}\Vert(\mathbf{W}\mathbf{x} + \mathbf{b}) - (\mathbf{W}\mathbf{0} + \mathbf{b})\Vert_2^2 = \frac{1}{2}\Vert \mathbf{W}\mathbf{x} \Vert_2^2$, and the derivative of this equation is the desired result.

% \subsection{Proof of Theorem 1.}

\subsection{Computing Exact the Lipschitz Constant of Networks is NP-hard}

We take a 2-layer fully-connected neural network with ReLU activation function as an example to demonstrate that Lipschitz computation is not achievable in polynomial time. As we denoted in Method Section, this 2-layer fully-connected neural network can be represented as
\begin{equation}
     f(\mathbf{W}^1,\mathbf{W}^2;\mathbf{x})  = (\mathbf{W}^{2}\circ\sigma \circ\mathbf{W}^{1})(\mathbf{x}),
\end{equation}
where $\mathbf{W}^1\in\mathbb{R}^{d_0\times d_1}$ and $\mathbf{W}^2\in\mathbb{R}^{d_1\times d_2}$ are matrices of first and second layers of neural network, and $\sigma(x)=\max\{0, x\}$ is the ReLU activation function.

\noindent\textbf{Proof.} 
To prove that computing the exact Lipschitz constant of Networks is NP-hard, we only need to prove that deciding if the Lipschitz constant $\Vert f\Vert_{Lip} \leq L$ is NP-hard.

From a clearly NP-hard problem:
\begin{align}
     \label{eq:nphard}
     \max\min & \Sigma_i (\mathbf{h}_i^{\mathsf{T}} \mathbf{p})^2 = \mathbf{p}^{\mathsf{T}} \mathbf{H} \mathbf{p}\\
    & s.t.  \quad  \forall k, 0\leq p_k\leq1,
\end{align}
where matrix $\mathbf{H}=\Sigma_i \mathbf{h}_i \mathbf{h}_i^{\mathsf{T}}$ is positive semi-definite with full rank.
We denote matrices $W_1$ and $W_2$ as
\begin{equation}
    \mathbf{W}_1 = (\mathbf{h}_1, \mathbf{h}_2,\cdots,\mathbf{h}_{d_1}),
\end{equation}
\begin{equation}
    \mathbf{W}_2 = (\mathbf{1}_{d_1\times 1},\mathbf{0}_{d_1\times d_2 -1})^{\mathsf{T}},
\end{equation}
so that we have
\begin{equation}
    \mathbf{W}_2 \textnormal{diag}\left(\mathbf{p}\right) \mathbf{W}_1 =  \begin{bmatrix}
    \mathbf{h}_1^{\mathsf{T}} \mathbf{p} & 0 &  \dots & 0  \\
    \vdots & \vdots & \ddots &   \\
    \mathbf{h}_n^{\mathsf{T}} \mathbf{p}  & 0 &  &  0
    \end{bmatrix}^{\mathsf{T}}
\end{equation}

The spectral norm of this 1-rank matrix is $\Sigma_i (\mathbf{h}_i^{\mathsf{T}} \mathbf{p})^2$. We prove that Eq. \ref{eq:nphard} is equivalent to the following optimization problem
\begin{align}
\label{eq:nphard2}
     \max\min & \Vert \mathbf{W}_2 \textnormal{diag}\left(\mathbf{p}\right) \mathbf{W}_1 \Vert_2^2 \\
    & s.t. \quad   \mathbf{p} \in \left[0, 1\right]^n.
\end{align}
Because $H$ is full rank, $W_1$ is surjective and all $\mathbf{p}$ are admissible values for
$\nabla g(\mathbf{x})$ which is the equality case. Finally, ReLU activation units take their derivative within $\{0,1\}$ and Eq. \ref{eq:nphard2} is its relaxed optimization problem, that has the same optimum points. So that our desired problem is NP-hard.

\end{document}